%% file: main.tex
\documentclass[pdflatex, sn-vancouver, 12pt]{sn-jnl}
\jyear{2023}

\usepackage{setspace}

\usepackage{geometry}
\begin{document}


\title[The Deep Generative Decoder]{The Deep Generative Decoder: \mbox{MAP estimation} of representations improves modeling of single-cell RNA data}

\author[1]{\fnm{Viktoria} \sur{Schuster}}\email{viktoria.schuster@sund.ku.dk}

\author*[1,2]{\fnm{Anders} \sur{Krogh}}\email{akrogh@di.ku.dk}

\affil[1]{\orgdiv{Center for Health Data Science}, \orgname{University of Copenhagen}, \orgaddress{\street{Blegdamsvej 3B}, \city{Copenhagen}, \postcode{2200}, \country{Denmark}}}
\affil[2]{\orgdiv{Department of Computer Science}, \orgname{University of Copenhagen}, \orgaddress{\street{Universitetsparken 5}, \city{Copenhagen}, \postcode{2100}, \country{Denmark}}}

\abstract{
\textbf{Motivation:} Learning low-dimensional representations of single-cell transcriptomics has become instrumental to its downstream analysis. The state of the art is currently represented by neural network models such as variational autoencoders (VAEs) which use a variational approximation of the likelihood for inference.\\
\textbf{Results:} We here present the Deep Generative Decoder (DGD), a simple generative model that computes model parameters and representations directly via maximum a posteriori (MAP) estimation. The DGD handles complex parameterized latent distributions naturally unlike VAEs which typically use a fixed Gaussian distribution, because of the complexity of adding other types. We first show its general functionality 
on a commonly used benchmark set, Fashion-MNIST. Secondly, we apply the model to multiple single-cell data sets. Here the DGD learns low-dimensional, meaningful and well-structured latent representations with sub-clustering beyond the provided labels. The advantages of this approach are its simplicity and its capability to provide representations of much smaller dimensionality than a comparable VAE.\\
\textbf{Availability and implementation:} The code is made available in \href{https://github.com/Center-for-Health-Data-Science/DGD_paper}{this} GitHub repository. scDGD is available as a python package in \href{https://github.com/Center-for-Health-Data-Science/scDGD}{this} repository.\\
\textbf{Contact:} akrogh@di.ku.dk\\
\textbf{Supplementary information:} Supplementary data are available at Bioinformatics online.\\
}


\singlespacing
\maketitle


\section{Introduction}

High-throughput methods in biology and medicine produce vast amounts of high-dimensional noisy data that we seek to extract knowledge from. The first step is often to obtain a low-dimensional representation of the data through clustering, PCA analysis or other similar techniques. A good example is gene expression analysis, where we need to compare counts for each of the 10-20 thousand genes between samples or cells. Here the first analysis is often to visualize the data in two dimensions with UMAP \cite{McInnesEtal2018} or t-SNE \cite{vanderMaaten2008}. Although these tools are indispensable, they do not model the uncertainty of the count data or reveal any of the underlying structure in the data. Therefore, it has become popular to make use of generative models that give low-dimensional representations mapping to a probability distribution over the data. Some of the most common are variational autoencoders (VAEs) \cite{Kingma2013}, generative adversarial networks (GANs) \cite{GoodfellowEtal2014}, normalizing flows \cite{Rezende2015} and energy-based models (EBMs) \cite{ebm_lecun}. Even though these approaches have their shortcomings \cite{bondtaylor2021deep}, they have shown great success in a multitude of applications such as single-cell RNA-sequencing (scRNA-seq) \cite{lopez_deep_2018,seninge_vega_2021}, image and surface generation \cite{abukmeil_survey_2021,kammoun_generative_2022}, and natural language processing \cite{abukmeil_survey_2021,wali_generative_2022}. In addition, diffusion models \cite{ho_denoising_2020} and Transformers \cite{vaswani_attention_2017} are among the best performing and most popular approaches in generative modeling of image and text data. However, these methodologies do not necessarily include probabilistic latent spaces.

The VAE is the most common type of generative model applied to biological data. In general, VAEs consist of an encoder that maps data to a lower-dimensional probabilistic representation in latent space and a decoder that maps back to the data. The model is trained using maximum likelihood estimation (MLE), which seeks to find a set of parameters that maximize the likelihood of the data. However, the likelihood calculation is intractable due to an integral over the latent space. The solution in the VAE is to approximate the likelihood and maximize a variational lower bound (the ``ELBO"). This approximation is not always accurate \cite{CremerElal2018} and favours larger latent dimensionalities than necessary \cite{yacoby_failures_nodate}.

The decoder alone specifies the generative model and the encoder of the VAE is just a convenient tool for finding representations, but it adds a whole new set of modelling choices to be made, which influence the inference gap and expressiveness \cite{CremerElal2018}. As a result, encoder-less representation learning has been suggested many times in different ways. There are simple, deterministic approaches that maximise the likelihood over representations of the training data and the decoder parameters at the same time \cite{Han2017,Bojanowski2018,vad_2019}. Here and in our work, the representations are treated like the other model parameters and since gradients can be derived exactly, the model can be trained using standard gradient descent. There are also other encoder-less generative models, such as Gaussian Process Latent Variable Models (GP-LVMs) \cite{lawrence_gaussian_2003} or PCA-based models \cite{collins_generalization_2001,mohamed_bayesian_2008,townes_feature_2019}. In their vanilla form, they have issues with scaling and limited model complexity and have not quite gained traction in the application to biological data compared to VAEs. However,  scalable versions of GP-LVMs have been presented \cite{ahmed_grandprix_2019,verma_robust_2020,lalchand_modelling_2022} and PCA-based methods have been applied for the joint modelling of multiple single-cell modalities \cite{mourragui_percolate_2023}. Scalability in these GP-LVM applications is achieved by applying sparse Gaussian processes and amortised inference \cite{ahmed_grandprix_2019,verma_robust_2020,lalchand_modelling_2022}. As a result, they are still limited to the same problems of variational inference \cite{CremerElal2018} and sensitivity of latent initialisation \cite{lawrence_gaussian_2003}. 
The standard VAE and many other models use a fixed Gaussian distribution over latent space. This may result in under-expressive representations, and it has been shown that learning the parameters of a Gaussian mixture leads to improved generalization and clustering \cite{dilokthanakul2017deep,9020116}. 
However, this makes the variational inference even more complex, as it has to be combined with score matching \cite{vahdat_2021}, or other approaches. Another alternative approach worth mentioning is the VQ-VAE \cite{van_den_oord_neural_2017} which learns discrete representations with more powerful priors than a standard Gaussian, but still requires an encoder.

In this work, we present a probabilistic formulation of a generative decoder using maximum a posteriori (MAP) estimation of all parameters. MAP estimation is the Bayesian analog of MLE and seeks to find the set of parameters that maximize the posterior, which is the probability of the representations and model parameters given the observed data. There is no intractable step in this estimation, so both model parameters and representations can be estimated directly. The model we are introducing consists of a decoder (or generator), such as a feed-forward neural network, and a distribution over representations with learnable parameters, which are optional. Estimating the representations of the training samples as well as the parameters is straight-forward and can be done by gradient descent as in a non-probabilistic formulation \cite{kroghschuster2021}. One of the main advantages of this approach over variational inference is the ease and flexibility with which distributions over representations can be learned. The advantage over GP-LVMs is that the approach is scalable. Our emphasis here is on a model with a parameterized distribution over representations. This is motivated by the intuition that representations are meaningful and are likely to group samples with different properties. We think of this approach as a marriage of generative models and manifold learning, such as UMAP. The beauty of the approach lies in its simplicity and utility. The decoder parameters, latent representations and latent distribution are estimated in the same way. The simplicity makes it easy to extend the model to more complex distributions and losses. Unlike in Gaussian Mixture VAEs \cite{dilokthanakul_deep_2017,bai_gaussian_2022}, no additional encoder for the Gaussian components is needed. It also makes it much simpler to understand for non-specialists. This is especially important, as we see great potential for this approach in fields like biology and medicine.

In the following sections we present the theory of the model and experiments on two very different types of data. We first demonstrate the general functionality of the proposed model on the Fashion-MNIST \cite{xiao2017/online} benchmark dataset. This is a typical choice in deep learning and an important step, because it enables us to investigate and understand the model's generative capabilities and potential caveats. We further explore the complexity of the distribution over representations and compare latent space and generative capabilities to models using variational inference. We also showcase the extension of the approach to supervized learning in which the mixtures of the latent distribution represent specific data classes. After establishing the model's functionality, we show an application to single-cell gene expression count modelling. We chose a data set of peripheral blood mononuclear cells (PBMC) \cite{zheng_massively_2017}, for which we compare our approach in terms of data reconstruction and clustering performance to scVI \cite{Gayoso2022} and scVAE \cite{gronbech_scvae_2020}. The latter  is a variational autoencoder also using an adaptive mixture of Gaussians rather than a single fixed Gaussian to model the latent representation. We show that our model learns a representation and Gaussian mixture model which cluster well according to cell type with additional, previously unobserved sub-clustering. 
Finally we compare the two models on 11 different single-cell gene expression data sets using default settings of scVI and our model.

\section{Results}

\subsection{The model}

\begin{figure}
	\centering
	\includegraphics[width=0.9\linewidth]{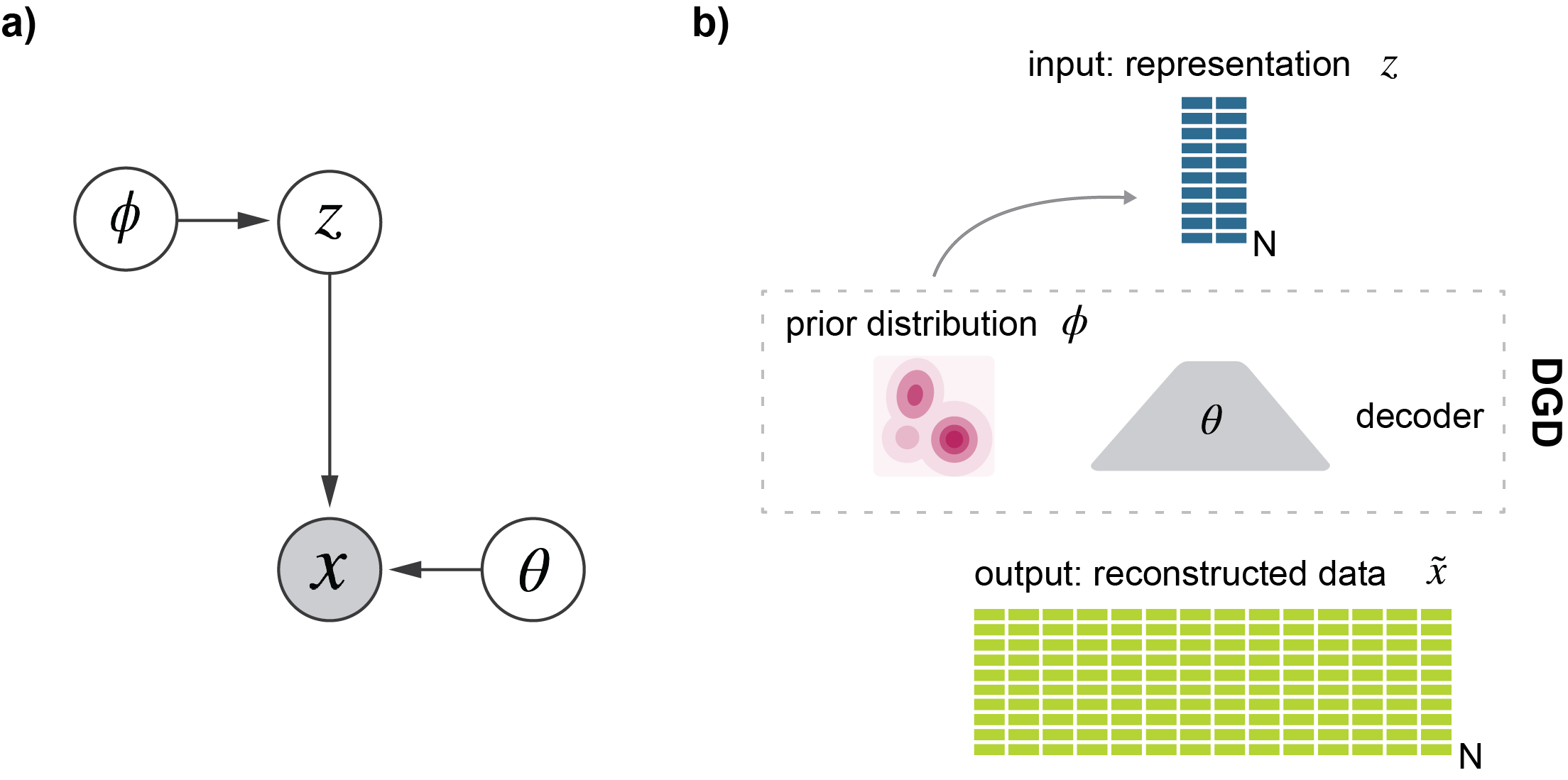}
	\caption{\textbf{The model.} \textbf{a)} Graphical model and \textbf{b)} schematic of the deep generative decoder. The DGD consists of a decoder of any desired architecture with parameters $\theta$ mapping the latent representation $Z$ to the data space $X$. The representation is modelled by a probability distribution with parameters $\phi$. $N$ is the number of samples.}
	\label{fig:dgd_scheme}
\end{figure}

Consider a model with observed variable $x$ and continuous latent variable $z$, which we also refer to as a representation. Usually, the $x$-space (or sample space) is of a higher dimension than the $z$-space, so the model gives a low-dimensional representation of data. A neural network, the decoder, with parameters $\theta$ takes $z$ as input and outputs $f_\theta (z)$, which are the parameters for the distribution of $x$. These could be the means and standard deviations of independent normal distributions. Thus, the neural network defines a conditional distribution in sample space, $P(x\mid z,\theta )= P(x\mid  f_\theta (z) )$. We assume a distribution over representation space, $P(z\mid \phi)$, with parameters $\phi$. The parameters may be fixed as in a standard VAE with a fixed Gaussian distribution over representations. It can also have adjustable parameters, such as a mixture of Gaussians with trainable means, covariances and mixture coefficients, but essentially any (differentiable) parameterized distribution can be used. When introducing priors over parameters, the joint probability of everything becomes

\begin{equation}
\begin{split}
P(x,z,\phi,\theta) & = P(x,z\mid \phi,\theta)P(\phi)P(\theta) = P(x\mid z,\theta)P(z\mid \phi) P(\theta) P(\phi).\label{eq:jointProb}
\end{split}
\end{equation}

Again, $P(x\mid z,\theta)$ represents the decoder neural network, $P(z\mid \phi)$ the distribution over representations, $P(\theta)$ the prior over the decoder parameters (neural network weights), and $P(\phi)$ the prior over the parameters in the representation space distribution. We use maximum a posteriori (MAP) estimation to find optimal parameters and representations. For a data set $X = \{\mathbf{x}^1,\mathbf{x}^2,\ldots , \mathbf{x}^N\}$ with $N$ samples we thus want to find representations ${Z = \{\mathbf{z}^1,\mathbf{z}^2,\ldots , \mathbf{z}^N\}}$ and parameters ${\phi,\theta}$ that maximize ${P(Z,\phi,\theta\mid X)}$:

\begin{equation}
\begin{split}
	\mathop{\mbox{argmax}}_{Z,\phi,\theta} P(Z,\phi,\theta\mid X) = \mathop{\mbox{argmax}}_{Z,\phi,\theta} P(X,Z,\phi,\theta)/P(X) = \mathop{\mbox{argmax}}_{Z,\phi,\theta} P(X,Z,\phi,\theta).
\end{split}
\label{jointProb}
\end{equation}

If we assume independence among training samples, the log of this joint probability can be obtained from (\ref{jointProb}), with $i$ indexing the samples

\begin{equation}
\begin{split}
\log  P(X,Z,\phi,\theta) & = \sum_i \left( \log P(\mathbf{x}^i\mid \mathbf{z}^i,\theta) + \log P(\mathbf{z}^i\mid \phi) \right) +\log P(\phi)+\log p(\theta).
\end{split}
\label{eq:likelihood}
\end{equation}

This is the quantity we want to maximise with respect to both  the representations ($z$) for all samples and the model parameters ($\phi,\theta$). The training process can be imagined like this: A representation $z$ for each training sample is initialized. This can be a random sample from a chosen distribution or a zero-valued vector. Representations are then passed through the decoder to give $P(x\mid z,\theta)$  and losses are computed from this and $P(z\mid \phi)$. All parameters $\theta$, $\phi$, and $z$ are then updated via back-propagation. The implementation is efficient and straightforward and compatible with standard modules and loss functions in deep learning frameworks, such as PyTorch. Once the model is estimated, prediction on new data points is done by first finding an optimal representation by maximizing $P(x\mid z,\theta)P(z\mid \phi)$ as above while keeping all other model parameters fixed, i.e., do gradient descent in $z$ alone.

In this work we use a parameterized Gaussian mixture for the representations ($P(z\mid \phi)$) with diagonal covariance matrix and a mollified Uniform (``softball prior") as prior over component means. Hereafter, we will refer to this Gaussian Mixture Model as GMM. We use a Dirichlet prior as a prior on the mixture weights and a Gaussian prior on the log-variances of the components. We use a weight decay in the training of the decoder, which corresponds to a Gaussian prior on the weights.

\subsection{Demonstration on image data}

It is customary to test generative models on image benchmarks, so we begin our experimental analysis by applying our model to the Fashion-MNIST data. It allows us to understand and evaluate model behaviour, latent space and most importantly sampling quality much better than more complex, biological data. The decoder used for the Fashion-MNIST models is based on the DCGAN \cite{radford2016unsupervised} generator architecture with modifications using common methods in image generation and is described in detail in the Methods section and Supplementary figure 3. We limited the design space \cite{radosavovic2020designing} to a latent dimension of 20, as interpolation can only occur in low-dimensional representations \cite{balestriero_learning_2021}.
The remaining parameters and how they were chosen are described in the Methods section. The model was trained for 500 epochs. 

\begin{figure*}[!t]
	\centering
	\includegraphics[width=0.9\linewidth]{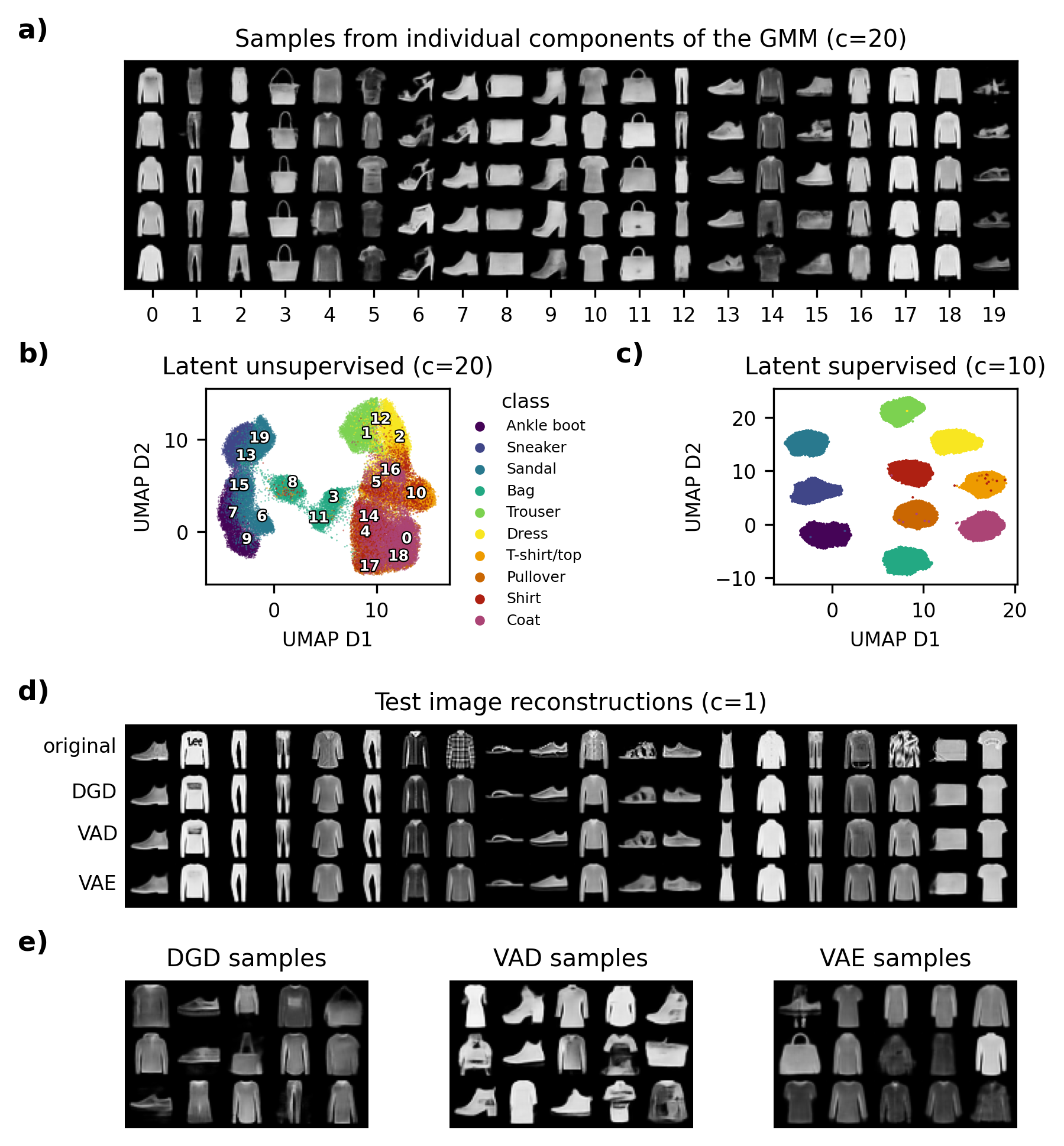}
	\caption{\textbf{Fashion-MNIST latent representations and samples.} The dimension of all latent spaces included in this figure is 20. The number of components of the GMM are denoted as ``c". \textbf{a)} Generated images of samples drawn from each component of the DGD with 20 Gaussian components. The number of the component drawn from is depicted on the x axis. 
	\textbf{b)-c)} Latent representations visualized as UMAP dimensions 1 and 2 colored by data classes. The UMAP was computed with 50 neighbors and a minimum distance of 0.7. \textbf{b)} Latent representation from the model with 20 GMM components. Numbers in the plot show the positions of the corresponding component means, from which samples were drawn in a). \textbf{c)} Latent representation from the supervised DGD with 10 components. \textbf{d)} Test image reconstruction from DGD, VAD and VAE. The top row shows the first 20 original test images. The remaining 3 rows show test image reconstructions. Names of the corresponding models are depicted on the left. The indication of one component refers to a standard Gaussian. \textbf{e)} Randomly sampled images from DGD, VAD and VAE. Corresponding models are indicated by plot titles.}
	\label{fig:structuredprior}
\end{figure*}

In figure \ref{fig:structuredprior}b) we see that a GMM with 20 Gaussian components distributes well over the complex latent space. The resulting distribution over latent space (the parameterized GMM) models the latent representation much better than for GMMs with less components which fail to pick up on more subtle structures (Supplementary figure 1). This supports the hypothesis stated in many works that overly simple distributions such as standard Gaussians may not be sufficient for describing the probabilistic representations of many data. Another advantage of the GMM is the increase in control over sampling. Figure \ref{fig:structuredprior}a) shows generated samples from individual components of the GMM. Having more components leads to each of them  covering a much smaller area of the latent space, and thus offering less varied but more specific samples. In this case of a 20-dimensional latent space with 20 mixture components, each component mean can be attributed to a (sub-)class of Fashion-MNIST. A relatively high number of components can thus enable more deliberate, more qualitative sampling and a better model of the latent space. While the latent representations in figure \ref{fig:structuredprior}b) show clear separation for super-classes such as shoes (Ankle boot, Sneaker, Sandal) and some  distinct classes like of trousers and bags, other classes are much more mixed. We evaluate the clustering with the adjusted Rand index (ARI) \cite{hubert_comparing_1985}. This metric achieves values between 0 and 1 with 1 representing a perfect clustering and is corrected for chance. The ARI of this clustering based on predicted labels derived from the GMM's component probabilities is only 0.295. As an extension of the DGD, it is also possible to perform supervised training, where each GMM component is assigned a data class a priori. In the optimization, each component is only committed to covering the latent space of its assigned classes' representations. When trained with these assigned 10 components, we arrive at the distinct latent clusters seen in figure \ref{fig:structuredprior}c) with an ARI of 0.995 (which is merely a sanity check, since the ARI is expected to be high). While the quality of random samples from this model is reduced slightly (see Supplementary figure 2) in comparison to those from an equivalent unsupervised model, we achieve perfect control over the sample identity.

We next investigated how the MAP approach compares to variational inference (VI) and amortization. For this purpose, we limit the prior of the DGD to a fixed standard Gaussian. We found that once the encoder of a VAE is removed (using the variational auto-Decoder (VAD) \cite{zadeh_variational_2021}), reconstruction performance improves significantly and is equal to that of the DGD, as seen in Figure \ref{fig:structuredprior}d) and Tables \ref{fmnist_model_comparison} and S1. Supplementary table 1 also shows that latent space normality is improved by removing the encoder. This supports \cite{kroghschuster2021,Bojanowski2018,bond-taylor_gradient_2021} in their results stating that the encoder can be a hindrance to achieving good representations. Given the standard Gaussian priors, the structure of the latent spaces is not practically changed between the three models (Supplementary figure \ref{fig:fmnist_umaps}), even though ARIs vary a lot (Supplementary table 1).

Our demonstration of the DGD on image data comes in very handy when next evaluating the generative modeling capabilities of the different approaches. The benefit of image data is that it comes naturally to us to evaluate whether generated data is sensible or not from a qualitative standpoint. In addition, there are many quantitative measures available. We make use of the Fréchet Inception Distance (FID) \cite{heusel_gans_2017}, a popular metric of the similarity between image distributions. It is, however, a biased metric depending on the number of samples and the generator itself. A qualitative analysis of the samples from all models (see figure \ref{fig:structuredprior}e)) ranks the DGD samples the highest in terms of realisticness and detail. However, the FID score of the VAE samples are slightly better (Table \ref{fmnist_model_comparison}) despite the generated images being very generic and blurry.

The FID score of the DGD improves drastically with learning a more complex parameterized distribution, which lends itself naturally to our approach. When using 20 Gaussian components, The FID decreases to 27.25\,$\pm$\,0.11 (from three computations using different random seeds, see the methods section), positioning itself between the probabilistic autoencoder (28.0) \cite{bohm_probabilistic_2022} and PeerGAN (21.73) \cite{wei_duelgan_2022}.

\begin{table*}
\centering
\caption{\textbf{Comparison of performance metrics for DGD, VAD and VAE with latent dimension 20 and one fixed standard Gaussian as latent prior.} Model names are indicated on the left. Performance metrics computed are indicated as the remaining columns. The BCE loss is the average binary cross-entropy loss of the reconstructed images, indicated with its standard error. The FID score is the Fréchet Inception Distance, a metric of the distance between generated and original image distributions. Both metrics are computed with respect to the test images. Reported means and errors stem from repeats with three different random seeds. We additionally report run times based on the same model architectures as an average over runs for three different latent dimensions (20, 50, 100) and 1000 epochs per run along with the standard error of the mean. The same hardware was used for all runs.}
\begin{tabular}{ p{2cm}p{2.5cm}p{2cm}p{3.5cm}}
\hline
Model & BCE loss & FID & Run time (sec/epoch)\\
\hline
DGD & \textbf{0.2629 $\pm$ 0.001} & 37.6 $\pm$ 0.2 & \textbf{25.2 $\pm$ 2.7}\\
VAD & \textbf{0.2615 $\pm$ 0.001} & 44.3 $\pm$ 0.2 & 35.7 $\pm$ 1.6\\
VAE & 0.2798 $\pm$ 0.001 & \textbf{33.9 $\pm$ 0.1} & 36.5 $\pm$ 10.3\\
\hline
\end{tabular}
\label{fmnist_model_comparison}
\end{table*}

\subsubsection{Application to single-cell data}

In order to assess the model performance on scRNA data, we chose to apply it to the peripheral blood mononuclear cells (PBMC) data from \cite{zheng_massively_2017}. The single cell expression counts are modeled by a Negative Binomial distribution. We achieved best clustering results with a latent dimension of 20 and 3 fully connected hidden layers. The model, which we will refer to as scDGD, was trained for 700 epochs after which it achieved its overall best performance in terms of reconstruction and clustering. The number of GMM components was set to 9, which represents the number of cell types in the data. Details about our implementation and how we arrived at these hyperparameters can be found in the Methods section and Supplements. 

\begin{figure*}[!t]
	\centering
        \includegraphics[width=0.9\linewidth]{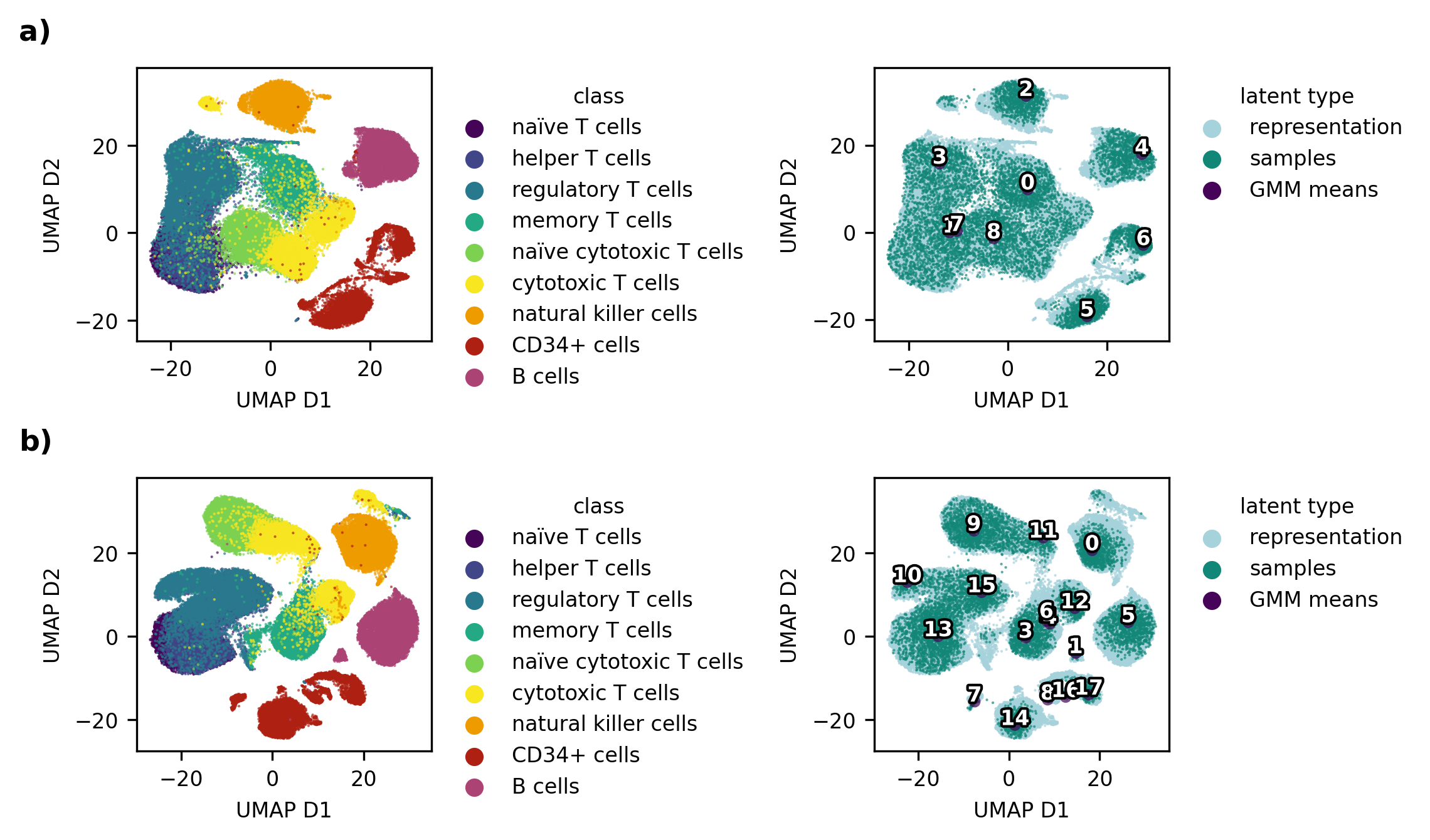}
        \includegraphics[width=0.9\linewidth]{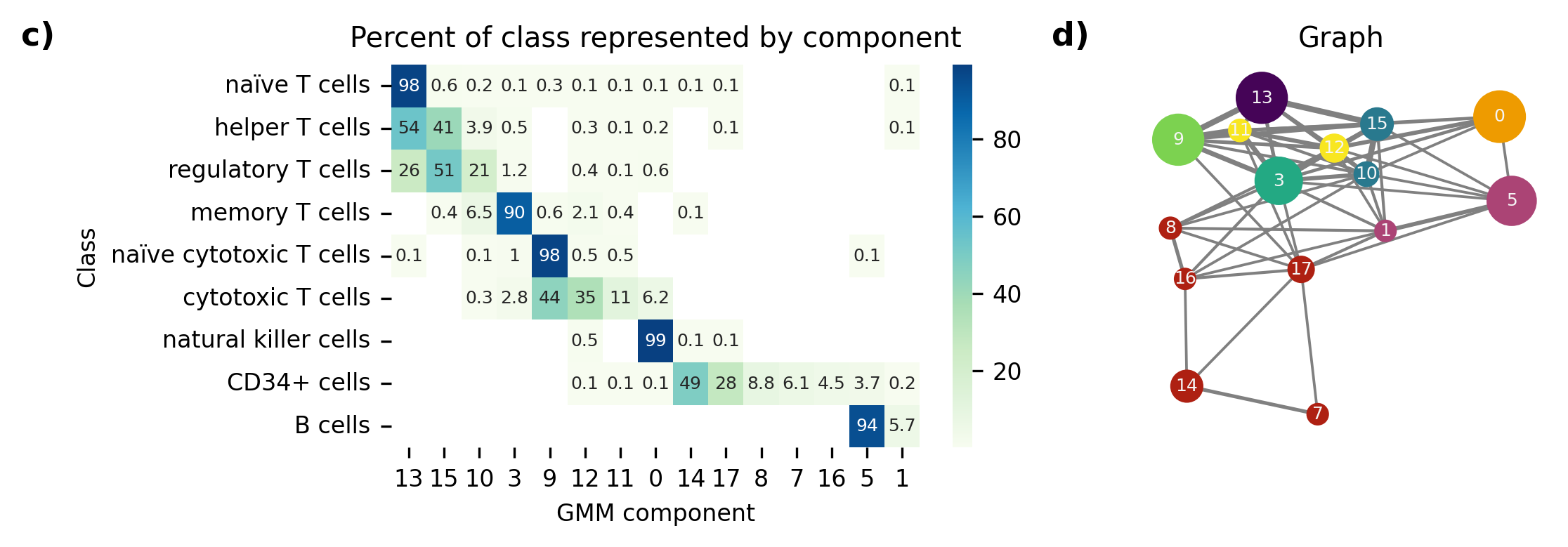}
	\caption{\textbf{Latent spaces of scDGD.} The latent spaces are shown colored by cell type (left) and type of latent point (right). Visualizations are achieved using UMAP with a spread of 5 and a minimum distance of 1. 'Representation' refers to learned representation of training data, 'samples' correspond to random samples drawn from the GMM and 'gmm means' represents the component means. The IDs of the component means are shown in black text on their corresponding coordinates. \textbf{a)} Latent space of scDGD with 9 Gaussian components. This model was trained for 700 epochs. \textbf{b)} Latent space of scDGD with 18 components, trained for 600 epochs. \textbf{c)-e)} Spatial relationships between GMM components of the 18-component scDGD. \textbf{c)} Heatmap of the GMM's component assignment to samples given as the percentage of each class with highest probability of a component. \textbf{d)} Graph visualization of the GMM component means with edge lengths correlated with Euclidean distances from a) and edge widths negatively correlated with  Euclidean distances. Only the 95th percentile of distances were accepted as graph edges, resulting in a threshold of 0.14. Edge colors are grey or show components belonging to the same cell type (same colors as in \ref{fig:pbmc1}b), except blue, referring to all CD4 cells minus memory T cells).}
	\label{fig:pbmc1}
\end{figure*}

The resulting latent space is depicted as a UMAP projection in figure \ref{fig:pbmc1}. 
From a visual perspective, the latent space and GMM components cluster well with the cell types. For clustering purposes, a cell is assigned to the component that gives the largest probability to its representation. The adjusted Rand index (ARI) of this clustering on the train set is $0.618$. 
In the work of \cite{gronbech_scvae_2020}, scVAE was reported to have a higher clustering performance than scVI of around 0.66 compared to 0.53 for scVI. However, scVI clearly outperformed the Gaussian Mixture VAE in terms of reconstructions. Since the scVAE tool could not be retrained, we as well compare our model to scVI and include the Leiden algorithm \cite{traag_louvain_2019} as a baseline for clustering performance. The clustering performance of scVI is evaluated based on Kmeans clustering of the latent space with 9 components. 

As a baseline for clustering performance, we apply the Leiden \cite{traag_louvain_2019} algorithm commonly used in the field of single-cell data. At 9 components, this achieves an ARI of $0.51$. A default scVI model trained on our train set resulted in a clustering performance of $0.566$, which is derived from Kmeans clustering with 9 components. The reconstruction performance of scVI is on par with our model trained with 9 components in terms of goodness of fit, but under-performs in reconstruction accuracy (see Table \ref{PBMC_model_comparison}). The reconstruction performance is evaluated based on two metrics. Firstly, we compute the negative log likelihood (NLL) of the true held out test counts being drawn from the gene-specific Negative Binomial distributions. These distributions are parameterized by the models' predicted means and learned dispersion parameters. Secondly, we also compute the root mean squared errors (RMSE) as a standard metric for comparability. We include both metrics for a more comprehensive understanding of the reconstruction performance. While the NLL measures the goodness of fit of the probabilistic modeling of counts and leaves more flexibility for genes that are highly variable, the RMSE emphasises the reconstruction accuracy and highlights sensitivities to outliers in the modeling. The clustering of scVI derived from the Kmeans algorithm with 9 components, however, is still lower than that of scDGD (see Table \ref{PBMC_model_comparison}). This shows that with scDGD, there is no need for a compromise with respect to built-in, high-performance clustering and count modeling.

\begin{table*}
\centering
\caption{\textbf{Comparison of performance metrics for scDGD and scVI.} Model names are indicated on the left, with c describing the number of Gaussians in the GMM (1 indicates a standard Gaussian). In the row for Leiden clustering, n/a denotes that the metric is not applicable for the given method. Performance metrics computed are indicated as the remaining columns. The NLL refers to the negative log-likelihood of the negative binomial distribution which models the counts in all methods. The best values for each metric are highlighted in bold.}
\begin{tabular}{ p{2.3cm}p{1cm}p{2.45cm}p{2.65cm}p{1.5cm}}
\hline
Model & ARI & NLL & RMSE & Run~time (min)\\
\hline
scDGD (c=9) & 0.618 & 2053.76\,$\pm$\,8.58 & \textbf{0.2246\,$\pm$\,0.0007} & 201.30\\
scDGD (c=18) & \textbf{0.684} & 2052.20\,$\pm$\,8.51 & \textbf{0.2249\,$\pm$\,0.0006} & 170.77\\
scVI (c=1) & 0.566 & 2053.99\,$\pm$\,8.52 & 0.2305\,$\pm$\,0.0007 & 200.05\\
Leiden & 0.512 & n/a & n/a & n/a \\
\hline
\end{tabular}
\label{PBMC_model_comparison}
\end{table*}

We also trained a model with 18 Gaussian components, since we have learned in our explorations on image data that increasing the complexity of the distribution over representation can be beneficial to modeling the substructures in the representation if the problem is complex enough. Increasing the number of components resulted in equal reconstruction performance as seen in Table \ref{PBMC_model_comparison} and highly improved clustering performance with an ARI of $0.684$. 
As we can see in figure \ref{fig:pbmc1}c), several cell types have one or more components specifically assigned to them. Among these are CD4+ naïve and memory T cells, NK, CD34+ and B cells. The cytotoxic T cell sub-types cannot easily be differentiated by component assignment, but are distinct from all other cell types in the data. The same goes for all CD4+ T cells except memory cells. Since t-SNE distorts the latent space, we also show a graph of the component means in Figure \ref{fig:pbmc1}d) based on the Euclidean distances between them (Supplementary figure 5).

\begin{figure*}[!t]
	\centering
	\includegraphics[width=0.9\linewidth]{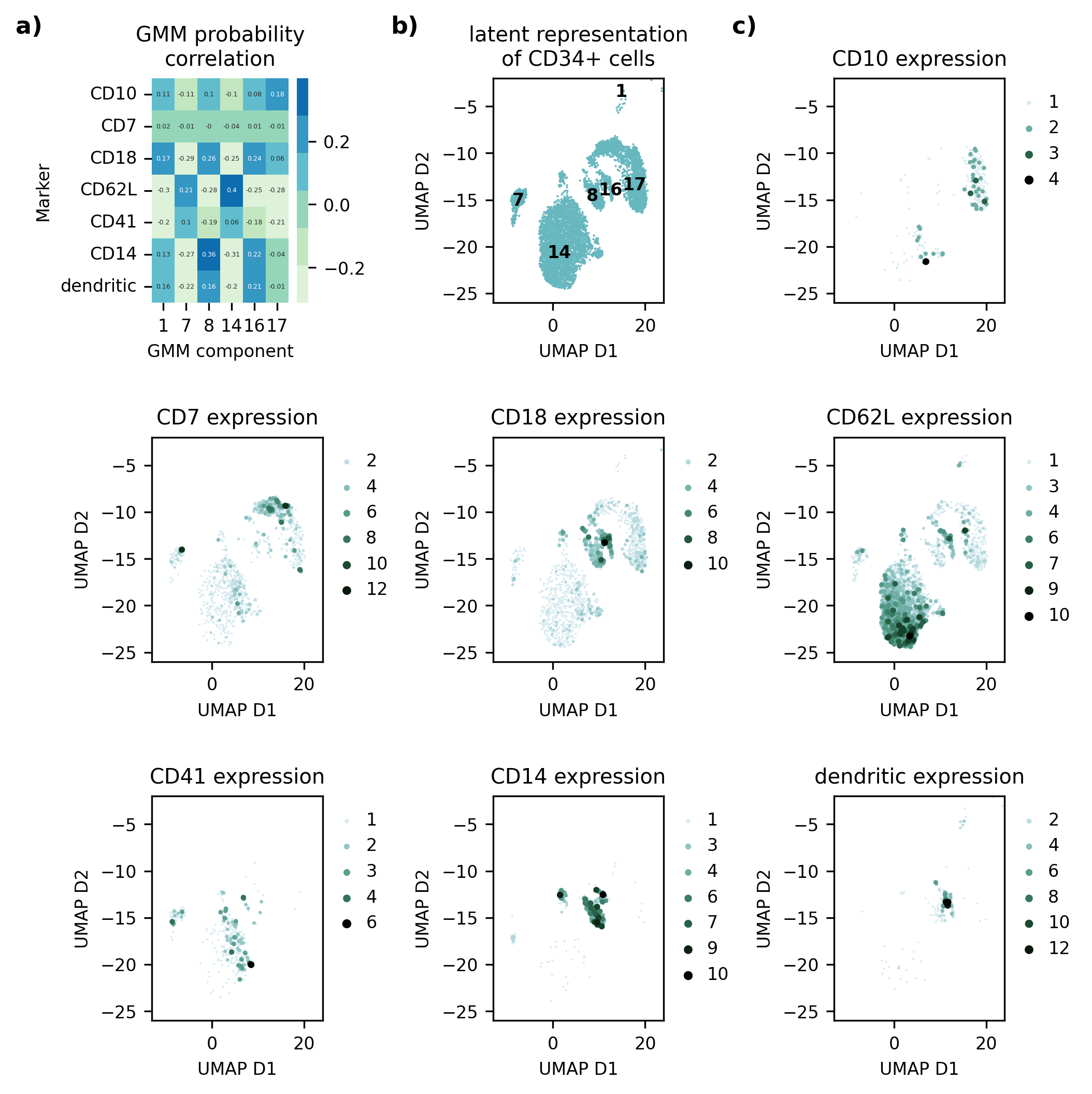}
	\caption{\textbf{UMAP projection of CD34-positive cell latent representations learned by scDGD with 18 Gaussian components coloured by expression levels of different genes.} \textbf{a)} Heatmap of Spearman correlation coefficients between GMM component probabilities and marker expression counts for all CD34-positive samples. \textbf{b)} Representations (green) and component means (white numbers) in UMAP projection of all CD34-positive cells on black background.  \textbf{c)} UMAP representation projection colored by expression counts for gene markers indicated by the plot titles.}
	\label{fig:cd34}
\end{figure*}

To go one step further in the analysis of our model, we next aimed to get an understanding of the structuring of the learned latent space. In both models with 9 and 18 components, we see that the CD34-positive cells form sub-clusters that are  modelled by different Gaussian components. We therefore investigated the representations for CD34-positive cells learned by the 18-component scDGD in terms of some more fine-grained differentiation markers. We were able to determine a sub-cluster best modeled by component 17 (see \ref{fig:cd34}a)) which is primarily occupied by cells expressing markers for the lymphoid lineage of hematopoetic stem cells (HSCs). This cluster further shows a distinction between bone marrow common lymphoid progenitors (CLPs) and cord blood CLPs, indicated by markers CD10 and CD7, respectively. Another sub-cluster, represented by components 8 and 16, is made up of several cells expressing CD18, which is a marker for myeloid cells regulating neutrophil production and a general marker for granulocytes. There is also a strong presence of CD14-positive cells in these clusters, suggesting an aggregation of monocytes and neutrophils. The biggest sub-cluster, modeled by component 14, suggests the presence of different sub-populations of the myeloid lineage of HSCs. We find gradual levels of CD41 to CD62L, which are markers for megakaryoblasts and myeloblasts, respectively. These are also correlated with component 7. We additionally found a concentrated site showing expression of different dendritic cell markers (CD1c, CD141, CD303), best represented by components 8 and 16.

\begin{figure*}[!t]
	\centering
	\includegraphics[width=0.9\linewidth]{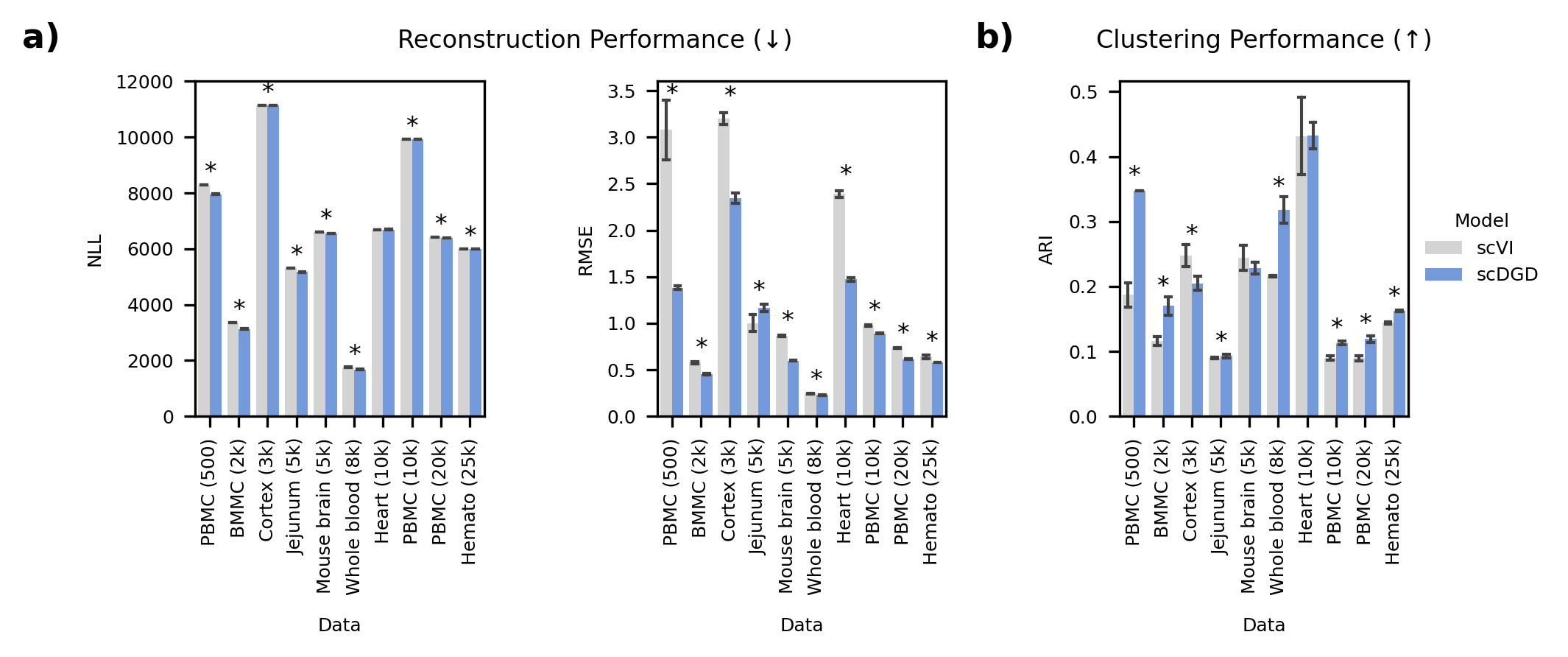}
	\caption{\textbf{Performance comparison of scVI and scDGD on independent data sets.} Error bars indicate the standard error of the mean from three models trained with different random seeds. Asterisks indicate whether differences are significant based on an alpha value of 0.05. \textbf{a)} Reconstruction performance measured as the negative log likelihood (NLL) of the negative binomial distribution and the root mean squared errors (RMSE). \textbf{b)} Clustering performance measured by the adjusted Rand index (ARI).}
	\label{fig:newdata}
\end{figure*}

Lastly, we investigated the merits of scDGD in terms of user experience and applied both default scDGD and default scVI to 10 data sets that had no part in the development of scDGD. Details about these data sets from mouse and human and their preprocessing can be found in the Methods section. Figure \ref{fig:newdata} shows that scDGD outperforms scVI in both data reconstruction (accuracy measured by RMSE) and latent clustering (in 9 out of 10 and 7 out of 10 cases, respectively). We also trained scVI and scDGD on the extremely large mouse brain data set with roughly 1.3 million cells \cite{}. On this large data set, scVI outperforms scDGD in terms of RMSE, but achieves again lower clustering performance and requires more computational resources (see Supplementary table \ref{mousebrain_model_comparison}).

\section{Discussion}

In this work, we have presented the DGD, a Bayesian formulation of a deep generative neural network using MAP estimation for model parameters and representations. This is a fully tractable and simple approach, for which no encoder is needed. This gives the DGD the advantage of having fewer parameters and fewer modelling choices, such as encoder architecture. We have shown that the DGD achieves good reconstruction, clustering and generative performance on much smaller latent spaces than some of the other generative approaches such as VAEs. In this sense it is comparable to flow-based models \cite{generativeLatentFlow}, but with less architectural restrictions. Additionally, we see clear advantages of this approach due to the simplicity of incorporating more complex distributions for modelling the latent representation. Analogous to our results on Fashion-MNIST, others have reported much better clustering and generative performances with a mixture of Gaussians compared to classic VAEs with standard Gaussians as priors over latent space \cite{gronbech_scvae_2020}.

In our application to single-cell expression counts, we showed that we can achieve reconstruction and clustering performances superior to those of comparable models, but with much smaller and better structured latent spaces, fewer parameters and more detailed distributions. Our choice for comparison are scVI \cite{Gayoso2022} and scVAE \cite{gronbech_scvae_2020}, two VAE-based methods 
modelling the representation of single-cell expression data. 
Overall, we compare scVI and our model on a total of 12 data sets, ranging from 500 to over one million cells. In these experiments, scDGD largely outperforms scVI on both reconstruction and clustering performance.
Investigating one of the learned representations closer, 
we find that the structure 
shows meaningful relationships and sub-clustering by Gaussian components, which can be linked to different sub-populations in some cell types. 

A caveat of this model is the inference time of new data points. The lack of an encoder means that the representations of new data points need to be found by back-propagation (with fixed model parameters). This process is more time-consuming and computationally expensive than passing new data points through an encoder as in the VAE-based model. However, in our experiments this only took 0.93 seconds to 36.24 minutes for 59 to 130613 test cells, respectively. 
In the case where inference is done repeatedly, one could also train an encoder afterwards as described in \cite{mapDGDv1}, which still results in a more expressive representation than the traditional AE setup.

Altogether, we find that the DGD is a simple and efficient approach to learning low-dimensional representations. Unlike some other generative models, it is fully tractable and bears no architectural restrictions. Although the model can be supervised as we showed on Fashion-MNIST, our scRNA model is completely unsupervised and is suitable for data without existing labels, while achieving a meaningful clustering of the representation on its own. We also observed that a more complex distribution relating to more GMM components can be beneficial for discovering more substructures in the data than provided by class labels. This makes the DGD an ideal candidate for modelling a multitude of biological and medical data. For these types of data, interpretability of the latent space is crucial. Several downstream applications such as perturbations and pseudo-time require a low-dimensional and continuous representation. We plan to expand the model to a variety of biological tasks and to include semi-supervised learning and other features that are of interest to generative models for biological and medical data. We further aim to apply this approach to even more extensive single-cell expression data sets and multi-omics data that allow a more thorough investigation of the latent space and can provide meaningful tools for data integration and downstream analysis.

\section{Methods} \label{methods}

\subsection{Data sets}

In this work we made use of in total 13 publicly available data sets, out of which 12 represent single cell transcriptomics sets. The data set referred to as PBMC was used to develop the application of the DGD to single cell expression data. The remaining single cell data sets were used to evaluate the performance of scDGD with default parameters in comparison to the popular and successful scVI model with default parameters.

\subsubsection{Fashion-MNIST}

The data set used in our proof of concept is the natural image Fashion-MNIST \cite{xiao2017/online} set. We used the data set's implemented train-test split. 

\subsubsection{PBMC}
The data set used to develop the single-cell application of our model is a single-cell gene expression count data set of peripheral blood mononuclear cells (PBMC) presented by \cite{zheng_massively_2017}. The data is provided by 10x Genomics under "Single Cell 3' Paper: Zheng et al. 2017 (v1 Chemistry)" and consists of data from the following 9 cell types: CD4+/CD45RA+/CD25- naïve T cells, CD4+ helper T cells, CD4+/CD25+ regulatory T cells, CD4+/CD45RO+ memory T cells, CD8+/CD45RA+ naïve cytotoxic T cells, CD8+ cytotoxic T cells, CD56+ natural killer cells, CD34+ cells, and CD19+ B cells. As \cite{gronbech_scvae_2020} we used the filtered gene-cell matrices. This data is extremely sparse with 98\% of the data being zero \cite{gronbech_scvae_2020}. Of the 32738 genes covered by this assay, only 21812 are expressed in the whole data set at least once. The 92043 samples are randomly split into train, validation and test set using percentages 81-9-10\%. The validation set is used to finding hyperparameters, and the test set is used for final evaluation.

\subsubsection{Single cell evaluation sets}

The following data sets were taken from 10x Genomics or chosen because of their use in the scVI paper \cite{Gayoso2022}. We used provided raw counts of cells that passed quality control filtering and did not apply any feature selection, modelling all transcripts available. Data sets were annotated using the CellTypist \cite{dominguez_conde_cross-tissue_2022,xu_automatic_2023} python package. The resulting cell type labels were used in the DGD for automatic selection of the number of Gaussian components and in the clustering performance evaluation as an approximated ground truth. All data sets were split into 80\,\% training, 10\,\% validation and 10\,\% test data. The raw counts with cell type and data split assignment as observables were exported as Anndata \cite{virshup_anndata_2021} objects which were used to train and evaluate both scDGD and scVI.

\paragraph{PBMC (500)}
The smallest of the data sets used for the elaborate analysis of our model's performance and applicability is another data set of PBMCs containing 587 samples and 36601 features from 10x Genomics\cite{pbmc500}. For cell type annotation, we used the `Immune\_All\_Low` model as reference and majority voting in CellTypist. This resulted in the presence of 11 distinct cell types.

\paragraph{BMMC (2k)}

The next data set contains 1985 bone marrow mononuclear cells (BMMCs) with 32738 features from 10x\cite{bmmc2k}. Cell type annotations were approximated using CellTpyist with the `Immune\_All\_Low` reference model and majority voting. This resulted in 15 distinct cell types.

\paragraph{Cortex (3k)}

A data set of 3005 cells from mouse cortex and hippocampus with 19972 features was published in \cite{cortex3k}. Cell type annotations were approximated using CellTpyist with the `Developing\_Mouse\_Brain` reference model and majority voting. This resulted in 14 distinct cell types.

\paragraph{Jejunum (5k)}

The 10x data set of human jejunum \cite{jejunum5k} contained 4392 cells with 36601 features. Cell type annotations were approximated using CellTpyist with the `Cells\_Intestinal\_Tract` reference model and majority voting. This resulted in 24 distinct cell types.

\paragraph{Mouse brain (5k)}

This data from 10x is comprised of 7377 cells from the adult mouse brain \cite{mousebrain5k} with 32285 features. Cell type annotations were approximated using CellTpyist with the `Developing\_Mouse\_Brain` reference model and majority voting. This resulted in 7 distinct cell types.

\paragraph{Whole blood (8k)}

From the blood of healthy females, a total of 8000 PBMCs, Neutrophils and Granulocytes \cite{wholeblood8k} were extracted by 10x, containing 36601 features. Cell type annotations were approximated using CellTpyist with the `Immune\_All\_Low` reference model and majority voting. This resulted in 10 distinct cell types.

\paragraph{Heart (10k)}

The 7713 cells with 31053 features from 10x in this data were extracted from an E18 mouse \cite{heart10k}. Cell type annotations were approximated using CellTpyist with the `Immune\_All\_Low` reference model. This resulted in 3 distinct cell types.

\paragraph{PBMC (10k)}

This data from 10x contains 11984 PBMCs from healthy female donors between the ages of 25 and 30, with 36601 transcripts covered \cite{pbmc10k}. Cell type annotations were approximated using CellTpyist with the `Immune\_All\_Low` reference model and majority voting. This resulted in 19 distinct cell types.

\paragraph{PBMC (20k)}

Another data set from female donors aged 25-30 provided by 10x was used, containing a total of 23837 cells with 36601 features. Cell type annotations were approximated using CellTpyist with the `Immune\_All\_Low` reference model and majority voting. This resulted in again 19 distinct cell types.

\paragraph{Hemato (25k)}

After excluding the \textit{basal-bm1} library due to poor quality (as presented in \cite{Gayoso2022} by recommendation of the authors), the data set of haematopoietic progenitor cells from mice \cite{hemato25k} was comprised of 25050 cells and 28205 features. Cell type annotations were approximated using CellTpyist with the `Cells\_Intestinal\_Tract` reference model and majority voting if cell types represented less than 100 cells. This resulted in 18 distinct cell types.

\paragraph{Mouse brain (1M)}

The extremely large data set of 1306127 brain cells from two E18 mice was provided by 10x with samples from cortex, hippocampus and the subventricular zone \cite{mousebrain1m}. It features 27998 transcripts. Cell type annotations were approximated using CellTpyist with the `Developing\_Mouse\_Brain` reference model and majority voting. This resulted in 27 distinct cell types.

\subsection{Models}

\subsubsection{The DGD}

The DGD consists of three components. Representation $Z$ is learned directly as a set of trainable parameters. For each of the $N$ samples of $X$, there exists an $m$-dimensional vector $\mathbf{z}$ that receives gradients. The values are initialized with zero for training but can be initialized with any other tensor of shape $(N,m)$. The decoder is characterized by the input dimension $m$ and output dimension $n$, which represents the number of data features. The decoder can be any type of Neural Network that fulfils this constraint. The choice of reconstruction loss is equally flexible but should ideally represent $P(\mathbf{x} \mid \mathbf{z})$. The distribution over latent space, which makes this a generative model, is here presented by a GMM. The mixture model consists of $K$ mixture components each with an $m$-dimensional mean vector $\pmb{\mu}$, and diagonal covariance $\mathbf{\Sigma}$ and a mixture coefficient $\mathbf{c}$ for each component. The mixture coefficients are transformed into component weights $\mathbf{w}$ through softmax activation. Since the diagonal covariance matrix can only take positive values, the according parameter is learned as the negative log-covariance.

The priors over the GMM parameters $\phi$ are as follows. The prior on the mixture weights is an $m$-dimensional Dirichlet distribution with uniform parameters, $\alpha$. The prior on mixture means is a mollified uniform which we call the "softball" distribution with hyperparameters scale as the radius of the $m$-ball and sharpness determining the slope of the boundary. Details on initialization and formulation of the log-probability can be found in Supplementary section ``The softball prior". 
For the negative log-variance, we use a Gaussian prior $\mathcal{N} (-2 \, \mathrm{log} (\sigma),1)$ with the same mean and standard deviation for all dimensions. Mixture coefficients and negative log-variances are initialized with default values $1$ (so that components weights are uniformly $K^{-1}$) and $-2 \, \mathrm{log} (\sigma)$ with $\sigma = 0.2 \times \mathrm{scale} \times K^{-1}$, respectively. The default values for $\alpha$, $\mathrm{scale}$, and $\mathrm{sharpness}$ are all $1$. Component means are initialized by sampling from the softball prior. The distributional component of the loss is given as 

\begin{equation}
\begin{split}
    - \, \mathrm{log} \, P(\mathbf{z}) = & - \, \mathrm{log} \, ( \, \sum_k \mathrm{exp} \, (\mathbf{w}^k \sum_i \mathrm{log}\,P(\mathbf{z}^i\mid \pmb{\mu}^k, \mathbf{c}^k, - \mathrm{log} \, \mathbf{\Sigma}^k))) - \, \mathrm{log}\,P(\pmb{\mu}, \mathbf{c}, - \mathrm{log} \, \mathbf{\Sigma})
\end{split}
\end{equation}

with $P(\mathbf{z}^i\mid \pmb{\mu}^k, \mathbf{c}^k, - \mathrm{log} \, \mathbf{\Sigma}^k)))$ as the density of the $k$th multivariate Gaussian component for representation $\mathbf{z}^i$. The training procedure is described in Algorithm \ref{alg:training}. Since representations are updated once per epoch and decoder and GMM parameters every batch pass, we use different instances of the optimizer for the different parameter sets. We have also observed that the GMM requires much larger steps than neural network parameters, so we choose to have an optimizer per parameter set with individually selected learning rates.

\begin{algorithm}
\caption{Training}
\begin{algorithmic}
\State Initialize parameters for representations $\mathbf{z}^i$, decoder and GMM \\
\FOR{epoch in n\_epochs:}\\
\hskip1em\FOR{$\mathbf{x}^i$, $i$ in training data:} \\
    \hskip2em $\mathbf{z}^i$ = $Z_{i.}$ \\
    \hskip2em $\mathbf{y}^i$ = model($\mathbf{z}^i$) \\
    \hskip2em loss = $L_{reconstruction}(\mathbf{y}^i, \mathbf{x}^i) + L_{GMM}(\mathbf{z}^i)$ \\
    \hskip2em Backpropagation \\
    \hskip2em Optimizer step for model and GMM \\
\hskip1em Optimizer step for Representation \\
\end{algorithmic}
\label{alg:training}
\end{algorithm}

The inference of new data points is straight-forward as well. For each new data point, one or more new representations are initialized. The type of initialization can be chosen. It can be from zero as done for the training data or from component means. For the latter initialization technique, the best starting point for each new sample is derived from the minimum reconstruction loss from all $K$ component means. This presents our default setting and essentially assigns the optimal component to each new sample. From there on, representations are typically optimized for a very short time (10 epochs) with a batch size of 32. Inference of new representations is of course done with frozen decoder and GMM parameters.

\subsubsection{Fashion-MNIST DGD}

For the image-generating DGD, we choose a network architecture based on convolutions and tricks from other image generation models and image segmentation techniques \cite{he_deep_2016,pmlr-v48-oord16,vahdat2020NVAE}. The decoders start with two fully connected hidden layers fed with the latent representations. The first hidden layer has 100 units, the second $capacity\times3\times3$ units. The capacity represents the minimum number of channels of the convolutional layers except for the gray-scale output channel and is set to 64. The output of the fully connected hidden layers is reshaped into $(batch,3,3,capacity)$ and fed into a series of NN blocks. These blocks consist of a transposed convolutional layer, swish activation and a Squeeze and Excitation layer as applied by \cite{vahdat2020NVAE}. Input- and output channels as well as kernel size, strike and padding depend on the position of the block in the scheme and can be seen in Supplementary Figure 3. There are four main blocks and two skip connection blocks. The outputs of combined blocks go through the activation function after summation. The series of these NN blocks is followed by a PixelCNN with 5 layers and mask size 5 \cite{pmlr-v48-oord16} and a last simple transposed convolutional layer reducing the number of channels to 1. The output is scaled using sigmoid activation.

Latent dimensionality and convolutional capacity vary depending on the experiment. For the models with 10 and 20 Gaussian components, we use a latent dimension of 20. The standard deviation of the GMM components are calculated as $\frac{scale}{components}$. This ensures that the components are sufficiently separated and alleviates the burden of another hyperparameter to optimize. The scale and hardness of the softball mean prior are 3 and 5, respectively, and the dirichlet alpha is set to 2 since we are dealing with balanced classes. For the models involving a standard Gaussian (including VAD and VAE), we test latent dimensionalities of 20, 50, and 100. Since the prior is not learned, softball and dirichlet prior are not relevant. The supplementaries contain results of a model with latent dimension 2, capacity 32 and varying number of Gaussian components (1, 10, and 20).

For training, we use the binary cross entropy (BCE) loss and Adam optimization with betas 0.5 and 0.7. For decoder, representation and GMM we choose learning rates $1e-3$, $1e-2$ and $1e-1$, respectively. This relationship has evolved as a rule of thumb for the DGD. Learning rates are best chosen with respect to a desired decoder learning rate (which is to be optimized for each task independently). From there, the representation learning rate should be ten times the decoder learning rate, and the GMM learning rate should be between 10 and 20 times that of the decoder.

Hyperparameters such as the number of hidden dimensions, capacity, dropout and the softball prior scale were found through optimization with respect to the validation reconstruction loss. The summary is available as a parallel coordinate plot from weights and biases runs in supplementary figure \ref{fig:fmnist_wandb}.

\subsubsection{VAD and VAE}

The VAD and VAE decoder implementations tested here are architecturally identical to the DGDs with a standard Gaussian. The VAE encoder was for simplicity chosen to mirror the decoder, with normal convolutional layers instead of the transposed convolutions. The capacity for all models is 32 and latent dimensions tested are 20, 50, and 100. Instead of the negative log density of the GMM, the loss for the distribution over latent space is given as the Kullback-Leibler divergence as typical for VI \cite{Kingma2013}.

\subsubsection{Supervised learning}

In the unsupervised model, the log likelihood $\log P(\textbf{z}^i\mid \phi)$ of an individual representation $\textbf{z}^i$ being drawn from the parameterized distribution $P(\phi)$ is given by the log sum of the probability densities of $\textbf{z}^i$ per component times the corresponding component probabilities, which are given by the softmax of the weights. In the supervised setting, we can ignore all components except the one that has been assigned to the sample's class. We thus only have to calculate the probability density of $\textbf{z}^i$ for the given component multiplied with the component probability. The losses for all other components will be zero and they will thus not receive gradients.

\subsubsection{Single-cell DGD (scDGD)}

The architecture of scDGD is much simpler than that of the Fashion-MNIST DGD. The decoder consists of the input layer with units defined by the latent dimensionality, and 3 hidden layers of each 100 units, connected to the output layer of (in the case of the PBMC data set) 32728 units, representing all transcripts in the data. The number of output units is given by the number of genes with non-zero expression counts in the data. The latent dimensionality was empirically found best to be 20 in terms of validation reconstruction loss and clustering accuracy of the cell types. The depth of the network was also empirically determined. The parallel coordinate summary of the hyperparameter optimization can be found in supplementary figure \ref{fig:pbmc_wandb}.

The normalized expression counts are modelled with a Negative Binomial distribution. Normalization is achieved by dividing the true expression count with the samples largest count. We can therefore use Sigmoid activation on the output layer. The reconstruction loss is given as the probability density of the re-scaled expression count and a gene-specific, learned dispersion parameter. Since this parameter is positive definite, we learn its logarithmic counterpart. For training, we use Adam optimization with betas 0.5 and 0.7. For decoder, representation and GMM we choose learning rates $1e-3$, $1e-2$ and $1e-2$, respectively. Two models are trained with 9 and 18 Gaussian components, respectively for the PBMC data set.

The component means of the GMM are distributed according to the mollified Uniform, for which we set scale and sharpness to $1$ each. This applies to both 9-component and 18-component model. The standard deviation of the components is initialized with a mean of $0.02$ for the 9-component model and $0.01$ for the 18-component model, which roughly corresponds to the formula we have introduced in the previous section, but is modified to a fifth of that in order to improve the component separation. Dirichlet alphas are set to $2$ and $1$ for models with 9 and 18 components, respectively. This is motivated by a dirichlet alpha of $1$ allowing for non-uniform component weights, which is desired in this case of imbalanced cell type classes. In the case of nine Gaussian components, a Dirichlet alpha of $1$ resulted in even more components representing T cells, so we decided to distribute the components more uniformly by increasing alpha.

All hyperparameters described above, except for the number of GMM components and of course the output dimensionality, present the empirically determined default parameters of scDGD. We applied this default model to 11 more data sets, in which the output dimensionality was set as the number of transcript found in the data, and the number of Gaussian components was determined by the number of unique cell types identified by CellTypist. Models for all data sets except PBMC (20k) were trained for 800 epochs. PBMC (20k) was trained for 1000 epochs. In all scDGD models, the learning rate of the decoder was reduced to $1e-4$ after 500 epochs.

\subsubsection{scVI}

An scVI model was trained on our PBMC train set. The model is implemented in Python version 3.8 with the scvi-tools \cite{Gayoso2022} package version 0.17.4. as described in \cite{lopez_deep_2018} with one hidden layer of 128 hidden units in both encoder and decoder. We chose a latent dimension of 20 for comparability with our model. The dropout rate is set to 0.1. The model was trained for 1000 epochs. For the analysis and comparison to scDGD, the latent space is clustered with Kmeans clustering (k=9) and then evaluated by the ARI. The lower bound is computed on our held out test set. Latent representations and ELBO were computed using scVI's implemented functions. Negative log likelihoods and RMSEs were computed based on the models returned mean and dispersion parameters for the test set.

For the remaining data sets, the scVI models were again initialized with default parameters and trained for up to 400 epochs, which represents the upper bound of the default number of epochs (for large data sets, the automatically calculated number of epochs can be very low).

\subsection{Performance metrics}

\subsubsection{Clustering metric}

We use the adjusted Rand index (ARI) \cite{hubert_comparing_1985} for evaluating and comparing clustering performance of our models. The value of the metric ranges from 0 to 1, representing random and identical clustering, respectively. This metric is relatively robust to different clustering approaches and diverging numbers of clusters, which makes it a good metric for comparing our method to others.

When analyzing models without GMM priors, the latent spaces are clustered using k-means \cite{kmeans} clustering implemented in Python's scikit-learn package.

\subsubsection{Reconstruction metrics}

For Fashion-MNIST, we use the binary cross-entropy (BCE) loss in order to evaluate the reconstruction of image pixels.

In scDGDs, we calculate the negative log likelihood as the summed log-density of parameterized Negative Binomials over all genes. For each gene, the log density is calculated based on the re-scaled model output and a gene-specific, learned, dispersion factor.

\subsubsection{Image generation evaluation metric}

For a quantitative evaluation of generated images, we employ the Fréchet Inception Distance (FID) \cite{heusel_gans_2017}. The FID gives the squared Wasserstein distance between two multivariate Gaussian distributions. We used the PyTorch implementation \cite{Seitzer2020FID} based on the original publication \cite{heusel_gans_2017}.
We compute the FID score three times for a given model, with random seeds 0, 21 and 87243. In each run, we randomly generate 10000 samples and compute the FID score with respect to the Fashion-MNIST test set.

\subsection{Software, statistics and visualization methods}

Python 3.8 is used as the programming basis for all methods. Neural Networks are implemented in Pytorch 1.10 \cite{NEURIPS2019_9015} and training progress was monitored and logged using \cite{wandb}. Dimensionality reduction is either done with our described model or common methods such as PCA or UMAP \cite{McInnesEtal2018}. If not stated otherwise, default parameters of 15 neighbors and a minimum distance of 0.1 are used in UMAP fits. Graph visualizations are achieved using the NetworkX package \cite{SciPyProceedings_11}. Requirements can be found in the corresponding repository \href{https://github.com/Center-for-Health-Data-Science/DGD_paper}{here}.

\subsection{Hardware}

Models were trained on NVIDIA TITAN Xp (12GB), except for scVI on the mousebrain (1M) data set, which was trained on NVIDIA TITAN RTX (24GB).

\section{Funding}
AK is supported by grants from the Novo Nordisk Foundation NNF20OC0062606, NNF20OC0059939, and NNF20OC0063268.

\section{Acknowledgments}
We acknowledge all the great discussions on representation learning and generative models with Wouter Boomsma, Jes Frellsen, Søren Hauberg, Aasa Faragen, Ole Winther and other members of Center for Basic Machine Learning Research in Life Science (MLLS), as well as support from our Center of Health Data Science on discussions about the methods and single-cell data (especially I\~{n}igo Prada Luengo) and for reviewing our manuscript (Jennifer Anne Bartell, I\~{n}igo Prada Luengo and Thilde Terkelsen). We especially thank Sarah Teichmann, her lab and Emma Dann for the insight we have gained into single-cell sequencing data which was invaluable in the reviewing process.

\section{Author contributions}

A.K. contributed the theoretical foundation of the approach. A.K. contributed to the implementation of the approach. A.K. contributed to the writing of the manuscript. V.S. contributed to the implementation of the approach. V.S. contributed to the writing of the manuscript. V.S. contributed the experiments. V.S. contributed the figures.

\section{Competing interests}
The authors declare no competing interests.

\section{Data availability}
No data sets were generated during the current study. All data used in this work is publicly available and referenced in the Methods section.

\section{Code availability}
The code is made available in \href{https://github.com/Center-for-Health-Data-Science/DGD_paper}{this} GitHub repository. scDGD is additionally available as a package in \href{https://github.com/Center-for-Health-Data-Science/scDGD}{this} repository.

\bibliographystyle{plain}
\bibliography{RepLearning2}

\setcounter{figure}{0}
\setcounter{table}{0}
\include{supplementaries}

\end{document}

%% file: supplementaries.tex


\subsection*{Supplementaries}


\subsubsection*{The softball prior} \label{softball}

In the softball prior used on the GMM means $\pmb{\mu}$, the means are initialized as 

\begin{equation}
\begin{split}
\pmb{\mu} & = \mathrm{scale} \times \mathbf{l} \times \frac{\mathbf{u}}{\| \mathbf{u} \|} \\
\mathrm{with} \, \, \mathbf{u} & \sim \mathcal{N}(0,1) \, \mathrm{and} \, \, \mathbf{l} \sim \mathcal{U}(0,1)^{m^{-1}}. \\
\end{split}
\end{equation}

\noindent The log-probability of this prior is given as 

\begin{equation}
\begin{split}
\mathrm{log} \, P(\pmb{\mu}) = A - \mathrm{log} \left(1+e^{\mathrm{sharpness}(\frac{\| \pmb{\mu} \|}{\mathrm{scale}} -1)}\right) \\
\mathrm{with} \, \, A = \mathrm{log} \, \Gamma(1+0.5m) - m (\mathrm{log} \, (\mathrm{scale}) + 0.5 \, \mathrm{log} \, \pi )\\
\end{split}
\end{equation}

\noindent The normalization constant can be understood as the volume of the $m$-ball and represents an approximation of the true normalization constant.

\subsubsection*{Figures and tables}

\begin{figure}[H]
	\centering
	\includegraphics[width=1.3\linewidth, angle=-90]{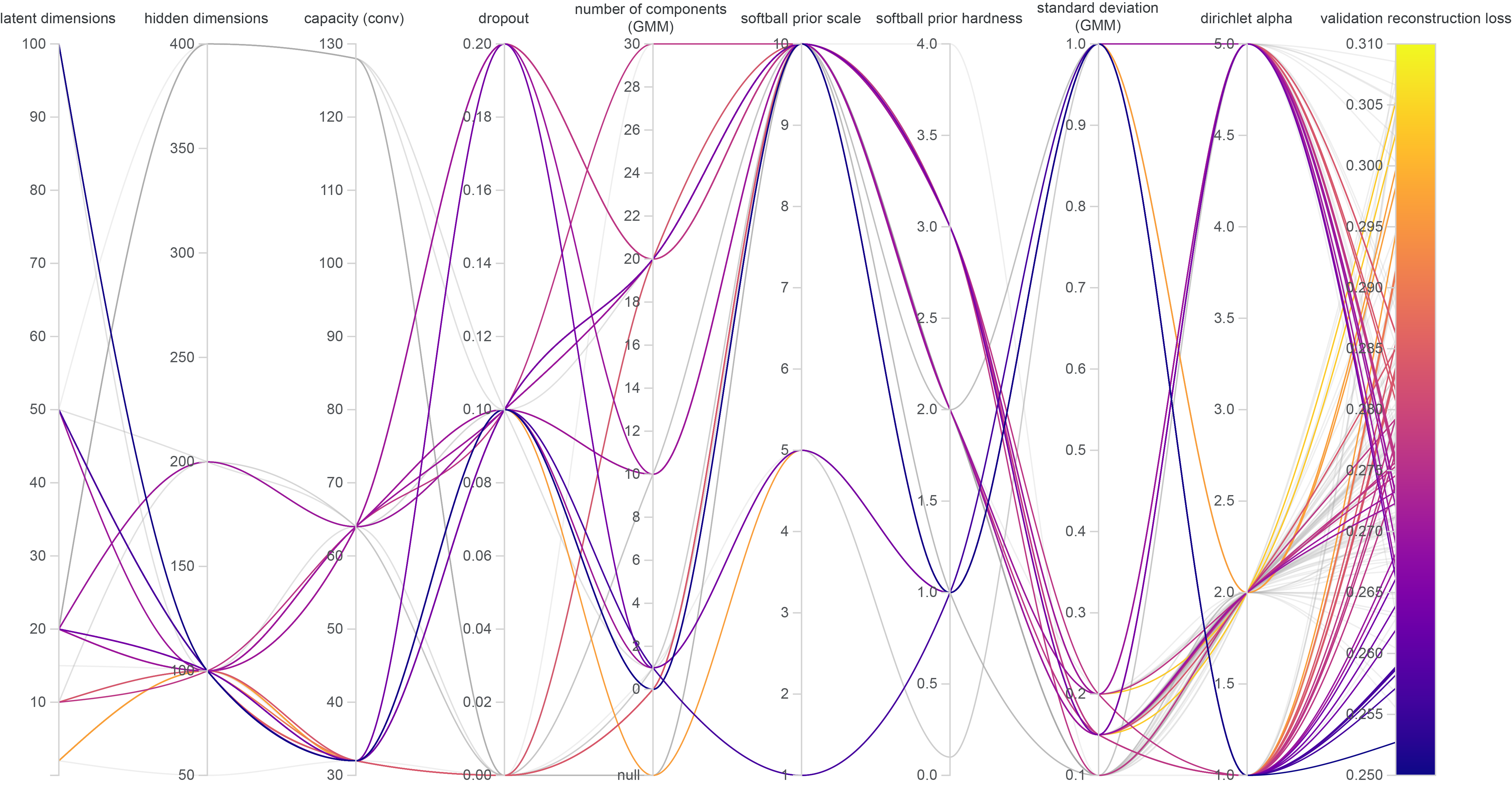}
        \caption{\textbf{Fashion-MNIST hyperparameter search.} The parallel coordinate plot from the corresponding wandb \cite{wandb} project shows combinations of hyperparameters (all coordinates of the plot except the last) and the resulting models' reconstruction performance on the validation set. Each model is represented by a line colored by the validation reconstruction loss (BCE). A total of 132 different models were tested.}
	\label{fig:fmnist_wandb}
\end{figure}

\begin{figure}[H]
	\centering
	\includegraphics[width=0.95\linewidth]{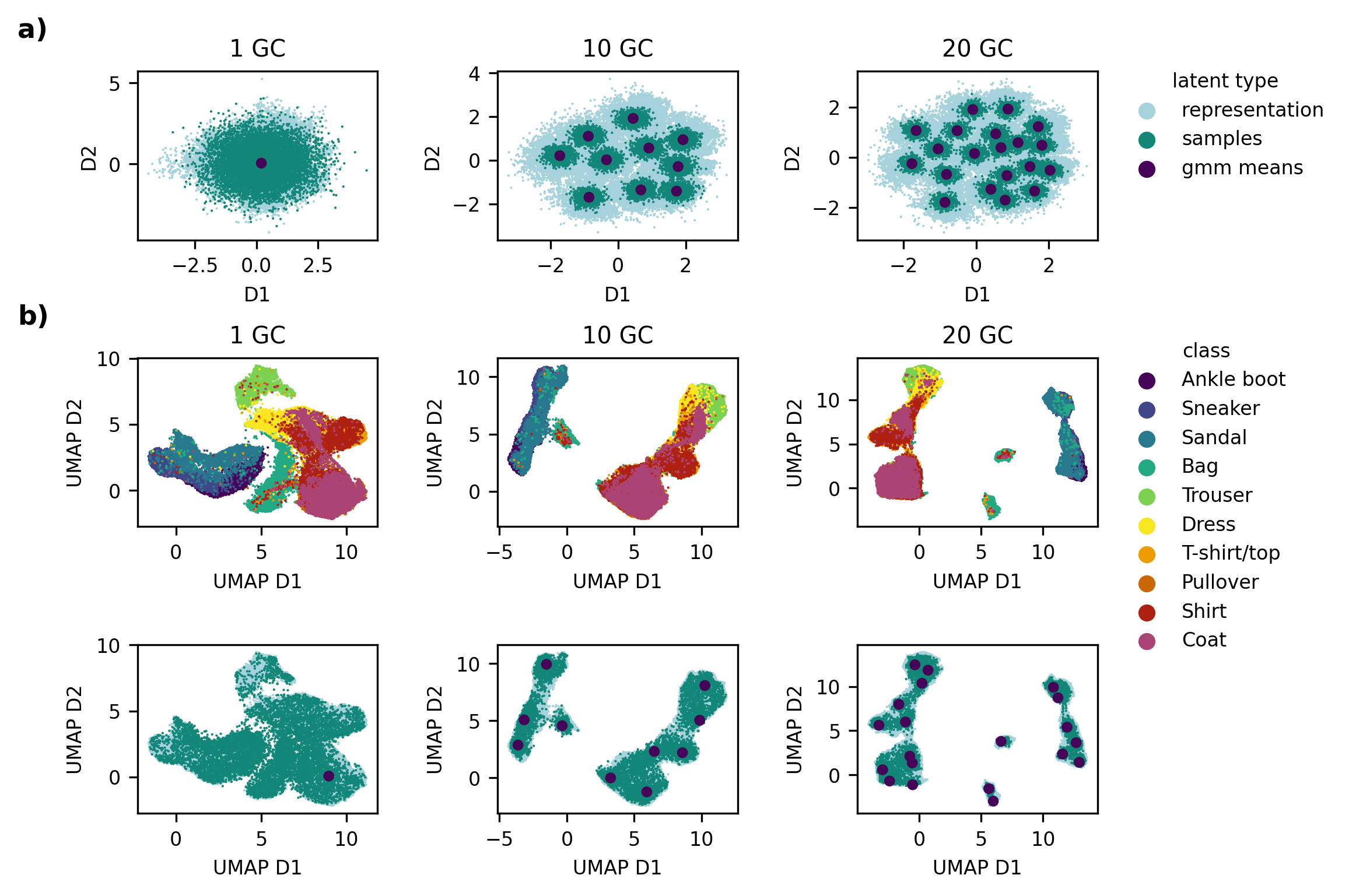}
	\caption{\textbf{Latent spaces for varying numbers of Gaussian components and latent dimensionalities trained on Fashion-MNIST.} \textbf{a)} DGDs with a 2-dimensional latent space are trained with 1, 10 and 20 Gaussian components (GC). Latent points are colored by their type. This refers to whether they are learned representations, samples drawn from the GMM or component means. \textbf{b)} UMAP projections of 20-dimensional latent spaces with 1, 10 and 20 Gaussian components. The top row is colored by sample class, the bottom by latent point type as in a).}
	\label{fig:structuredprior_supp}
\end{figure}

\begin{figure}
	\centering
	\includegraphics[width=0.95\linewidth]{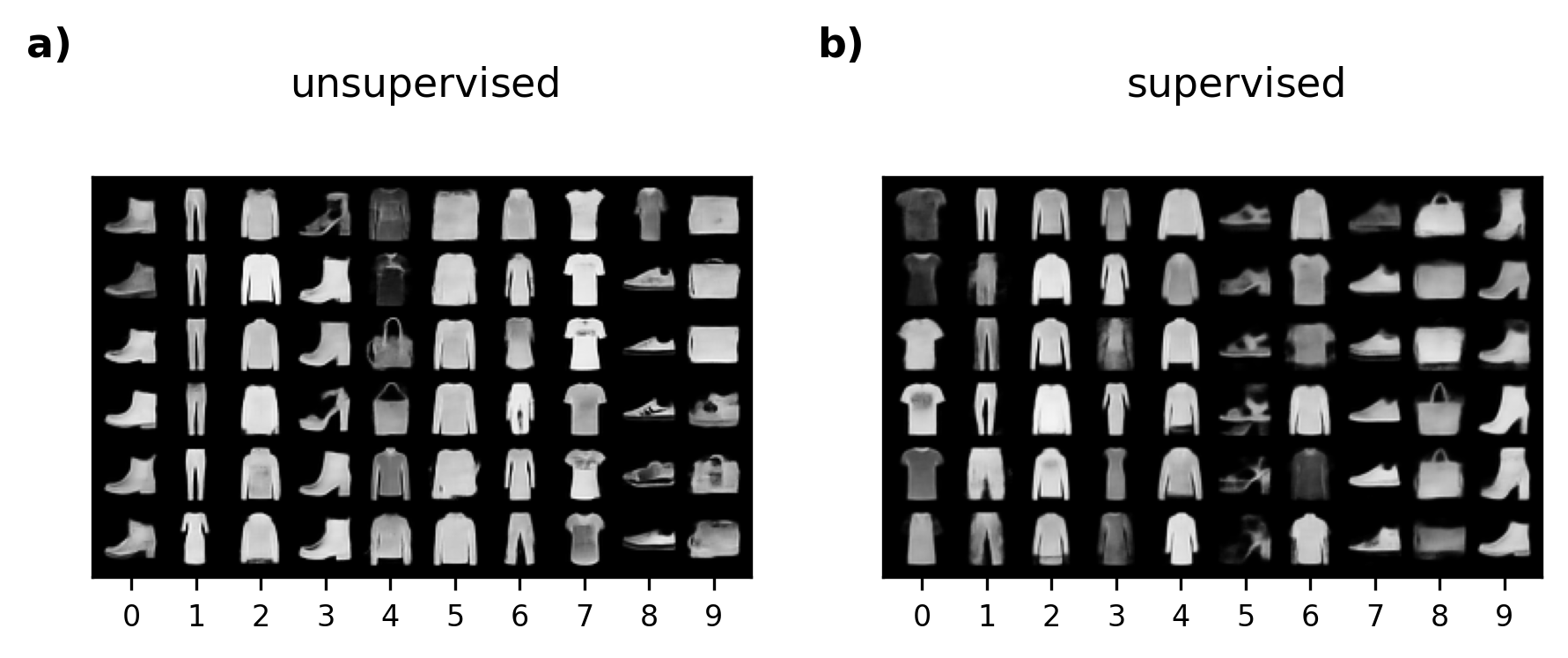}
	\caption{\textbf{Unsupervised and supervised learning of representation and GMM for Fashion-MNIST.} DGDs with a 20-dimensional latent space are trained with 10 Gaussian components in a \textbf{a)} unsupervised and \textbf{b)} supervised manner. For each model, 6 reconstructed samples are shown (row-wise) for each of the 10 components, indicated by the component ID below.}
	\label{fig:supervised_supp}
\end{figure}

\begin{table}
\centering
\begin{tabular}{ p{1.5cm}p{1cm}p{2.5cm}p{2cm}p{1.5cm}p{1.5cm} }
\hline
Model & Latent & BCE loss & Shapiro-Wilk test statistic & ARI & FID score\\
\hline
DGD & 20 & 0.2629 $\pm$ 0.001 & 0.9939 & 0.21 & 37.51\\
DGD & 50 & 0.2578 $\pm$ 0.009 & 0.9774 & 0.14 & 74.34\\
DGD & 100 & 0.2578 $\pm$ 0.009 & 0.9414 & 0.13 & 73.22\\
VAD & 20 & 0.2615 $\pm$ 0.001 &  0.9962 & 0.38 & 44.15\\
VAD & 50 & 0.2575 $\pm$ 0.009 & 0.9898 & 0.32 & 52.95\\
VAD & 100 & 0.2583 $\pm$ 0.001 & 0.9543 & 0.31 & 46.37\\
VAE & 20 & 0.2798 $\pm$ 0.001 & 0.7533 & 0.27 & 33.87\\
VAE & 50 & 0.2808 $\pm$ 0.001 & 0.4431 & 0.29 & 34.47\\
VAE & 100 & 0.2804 $\pm$ 0.001 & 0.2999 & 0.36 & 33.01\\
\hline
\end{tabular}
\caption{Metrics of the quantitative analysis of DGD, VAD and VAE trained and evaluated on Fashion-MNIST. The BCE loss represents the averaged Binary Cross-Entropy loss of the reconstructed images. The lower it is, the better. The Shapiro-Wilk test statistic describes the Normality of the latent representation. The test statistic lies within the range of 0 to 1, the higher the better.}
\label{MAP_vs_VI}
\end{table}

\begin{figure}[H]
	\centering
	\includegraphics[width=0.95\linewidth]{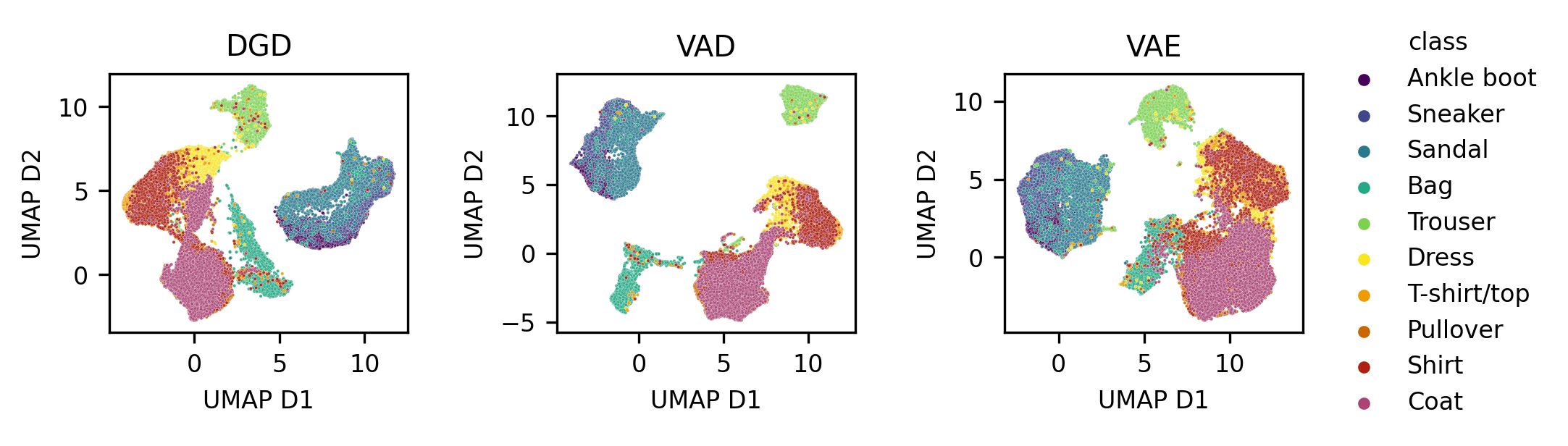}
        \caption{\textbf{Fashion-MNIST latent space visualizations of DGD, VAD and VAE train representations.} Representations are visualized in the first two UMAP dimensions and samples are colored by the class label.}
	\label{fig:fmnist_umaps}
\end{figure}

\begin{figure}
	\centering
	\includegraphics[width=1\linewidth]{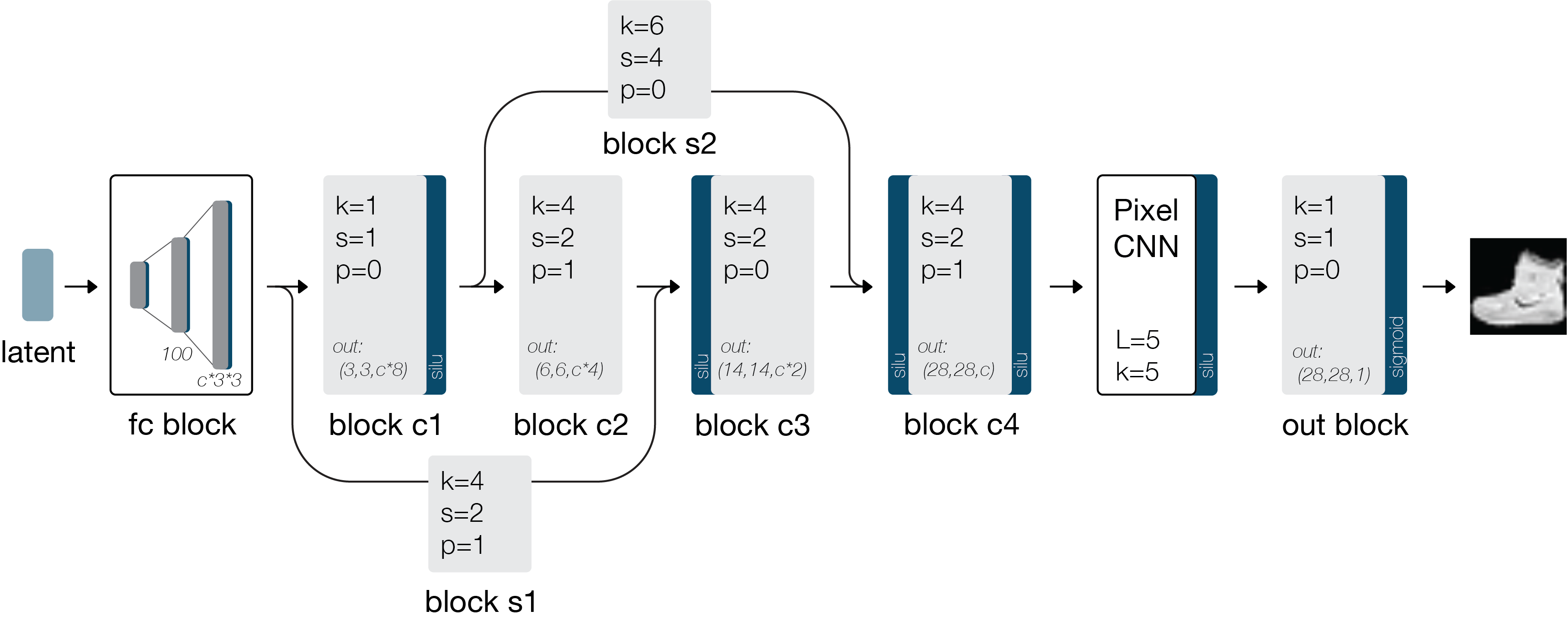}
	\caption{\textbf{Schematic of the Fashion-MNIST decoder architecture.} }
	\label{fig:fmnist_arch}
\end{figure}

\begin{figure}
	\centering
	\includegraphics[width=1.3\linewidth, angle=-90]{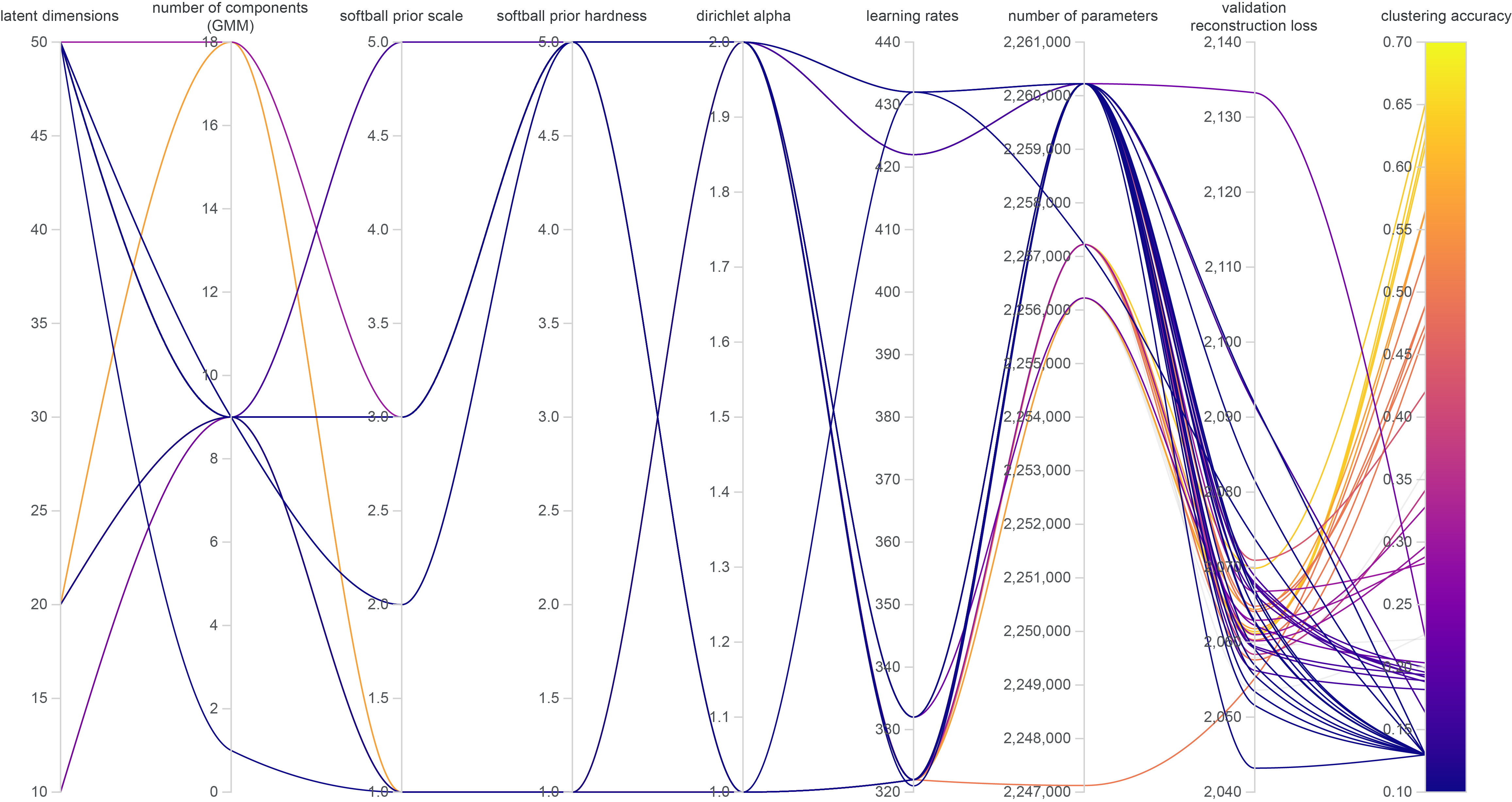}
        \caption{\textbf{Single-cell DGD hyperparameter search.} The parallel coordinate plot from the corresponding wandb \cite{wandb} project shows combinations of hyperparameters (all coordinates of the plot except the last) and the resulting models' reconstruction performance on the validation set, as well as the clustering accuracy. Each model is represented by a line colored by the clustering accuracy with respect to the cell types of the data. A total of 67 different models were tested.}
	\label{fig:pbmc_wandb}
\end{figure}


\begin{figure}
\centering
	\includegraphics[width=0.5\linewidth]{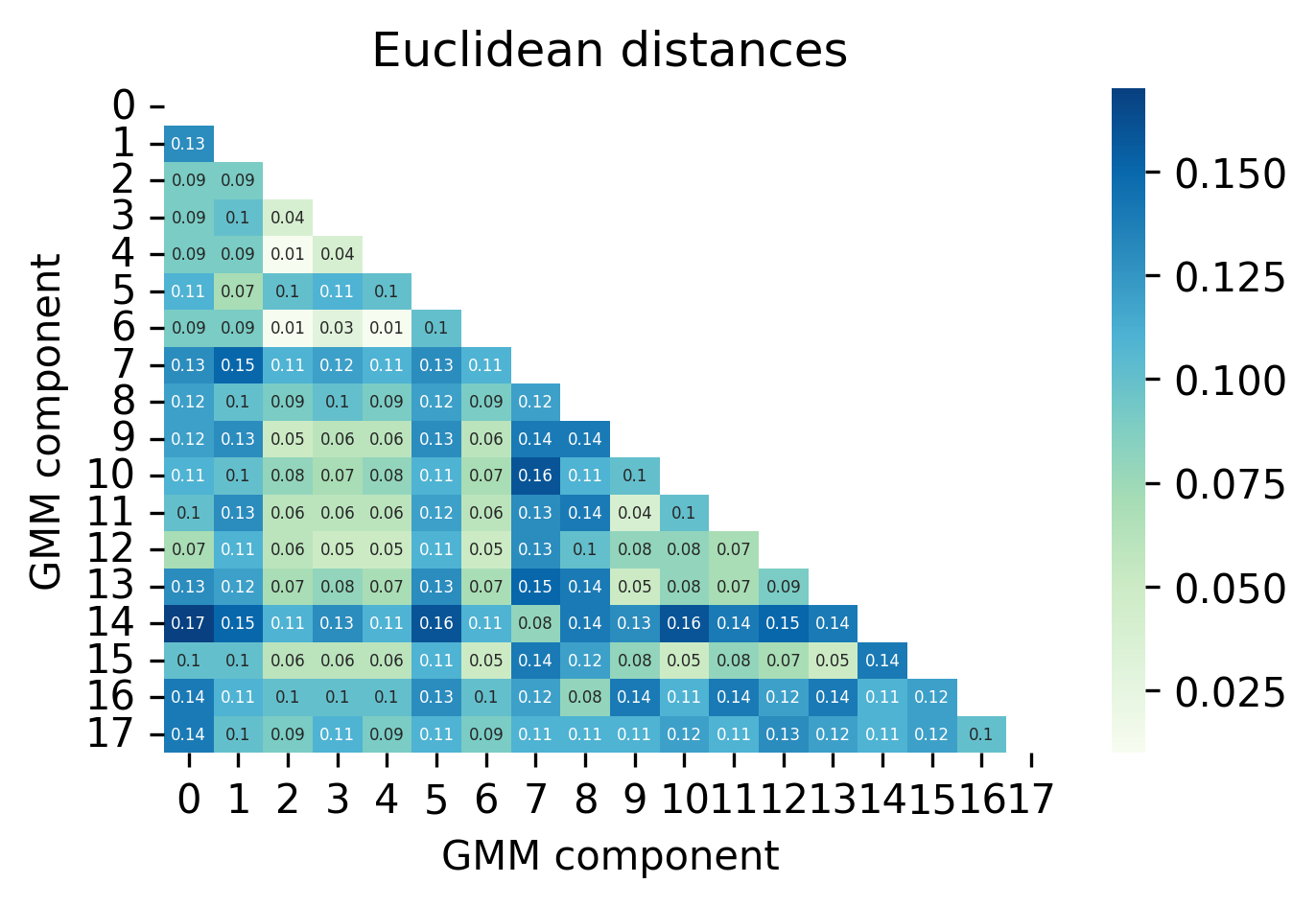}
	\caption{\textbf{Euclidean distances between GMM component means of the 18-component scDGD.} Heatmap of the Euclidean distances between GMM component means.}
	\label{fig:pbmc_components}
\end{figure}

\begin{table}
\centering
\caption{\textbf{Comparison of performance metrics for scDGD and scVI on the 1.3 million mouse brain cells data.} Model names are indicated on the left. Performance metrics computed are indicated as the remaining columns. The NLL refers to the negative log-likelihood of the negative binomial distribution which models the counts in all methods. The best values for each metric are highlighted in bold.}
\begin{tabular}{ p{1cm}p{1cm}p{1.3cm}p{1.3cm}p{1.3cm}p{1cm}p{2.2cm}}
\hline
Model & ARI & NLL & RMSE & Run~time (hours) & Epochs & Resources (max)\\
\hline
scDGD & \textbf{0.351} & 5382.60 $\pm$\,4.97 & 0.6153 $\pm$\,0.0035 & 25.82 & 800 & 53.1\,GB~(CPU), 9.1\,GB~(GPU)\\
scVI & 0.279 & 5377.91 $\pm$\,4.94 & \textbf{0.5622 $\pm$\,0.0021} & \textbf{13.81} & 400 & 124.5\,GB~(CPU), 1.7\,GB~(GPU)\\
\hline
\end{tabular}
\label{mousebrain_model_comparison}
\end{table}

